# The Linguistics Olympiads: Towards a New Corpus for Linguistics Research?

Vlad A. Neacșu
*"Iorgu Iordan – Al. Rosetti" Institute of Linguistics,*
*Romanian Academy*
vlad.neacsu2009@gmail.com

**Abstract**
Linguistics olympiad problems (LOPs) are a category of self-sufficient puzzles consisting of a scaled-down corpus representative of certain linguistic phenomena, from which the solver must deduce a primitive set of rules of the language and then translate a new set of elements. The linguistics olympiads (LOs) have become a worldwide phenomenon with 43 different territories taking part in the International Linguistics Olympiad (IOL) 2025. While the typology and solving strategies of LOPs have been analysed, their scientific facet and connections to academic linguistics have yet to be explored. LOPs are directly connected to many linguistic fields, e.g., linguistic typology, linguistic relativity, and linguistics fieldwork. Recently, LOPs have become a research focus as benchmarks for large language models, thus highlighting their usefulness in computational linguistics. Nevertheless, they have not yet been integrated into mainstream linguistics research.

This paper attempts to open new directions of including this particular type of puzzle in academic research by offering a structured evaluation of LOPs as linguistic data sources and proposes criteria for their responsible use in academic research. Starting from a set of over 1800 LOPs, this study critically examines the potential of LOPs as a novel corpus for linguistics research by discussing their strengths and limitations as tools, as well as the areas of linguistics into which these problems could fit. This work forms the foundation for a broader initiative aimed at bridging the gap between LOs and academic linguistics, by establishing a robust theoretical framework for LOPs.

## 1. Introduction

Linguistics olympiad problems (LOPs) are a special genre of puzzles or enigmas that present linguistic features and phenomena in an encrypted form (Derzhanski & Payne, 2010). They generally consist of a scaled-down corpus of sentences, phrases, or words representative of certain linguistic phenomena, based on which the solver can (1) infer and/or deduce a primitive set of linguistic patterns or rules (e.g., phonological, morphological, syntactic, semantic), and (2) apply the previously discovered rules to translate or construct a new set of sentences, phrases, or words.

The linguistics olympiad (LO) phenomenon was started in 1965 in the Soviet Union by Alfred Zhurinski in what was then called the "Traditional Olympiad in Linguistics and Mathematics". Since then, it has developed into a broad range of competitions at various levels, including the International Linguistics Olympiad (IOL), as well as numerous independent local, regional, and national linguistics olympiads (LOs). The main aim of the IOL and, by extension, the LOs is "to promote awareness of language, of the world's linguistic diversity, and of the essence of linguistics among pre-university students and the general public" (*Regulations of the International Linguistics Olympiad*, 2009). As such, these problems consisting of reduced-size corpora are meant to foreground the peculiarities and interesting features of languages, through self-directed and problem-based learning. The LOs have become a worldwide phenomenon with 43 countries and regions from five continents taking part in IOL 2025.

To this day, very little has been published about LOPs and especially about their integration into academic linguistics. The social practice of LOs was already thoroughly investigated by E. Cardoso Martins (2022), but the scientific facet of these problems and their implications and connections to academic linguistics is yet to be explored. A few articles were published about the linguistics competitions in Australia (Estival et al., 2014), Spain (Toral et al., 2014), and the United States (Radev et al., 2008), as well as some discussing the general linguistics competitions (Derzhanski & Payne, 2010).

Similarly, only a limited number of papers discuss the creation process of LOPs, but the discussion is mostly limited to their use in linguistics competitions for secondary school students (Bozhanov & Derzhanski, 2013; Derzhanski & Veneva, 2018; Iomdin et al., 2013; Littell et al., 2013; Nikulin, 2024; Verhoeven, 2016). Additionally, a few books have been published, containing collections of LOPs from national linguistics olympiads, for helping participants better train for this competition. They include the linguistics olympiads from the United States (Radev, 2013a, 2013b), Bulgaria (Derzhanski, 2009; Derzhanski & Velinov, 2012, 2013), Russia (Belikov et al., 2006), Romania (Rusu et al., 2017), and the Netherlands (Lubotsky & De Vaan, 2010). These collections contain the solutions to the problems, in most cases with a commentary discussing certain linguistics phenomena, but they are still addressed to secondary school students.

The usage of LOPs in secondary education in Brazil is explored by Costa & Silva (2022). This work discusses the way in which this type of problems can contribute to teaching foreign languages, as well as the way in which they can be used in the classroom.

My previous works (Neacșu, 2022, 2024) proposed a typological classification of LOPs into nine main types, broadly corresponding to the subfields of linguistics (grammatology, phonetics, phonology, morphology - noun and verb -, syntax, semantics, numeral systems, other types of problems). While previous works have also proposed classifications based on linguistic subfields, as extensively reviewed by Martins (2022, pp. 312–347), these works are among the first to propose a general approach to each type of problem, as well as general solving methods. Nevertheless, the target audience of this book is high school students preparing for the linguistics olympiads; therefore, a more rigorous and systematic scientific study is still needed.

## 2. Advantages of using LOPs in scientific research

In this section, I will discuss the main benefits of using LOPs as an unconventional corpus or collection for scientific research. These advantages make the LOPs a unique kind of corpus that can and should be used complementarily with traditional academic resources.

### *2.1. Linguistically-diverse corpus*

The main advantages of this corpus result from the aims of this genre of problems[1], to be used in student competitions, aiming for original compositions and non-repeating or similar problems, making this corpus linguistically diverse. Considering a sample of 1863 LOPs from national and international LOs, I identified 867 featured languages from 114 different macro language groups[2], totalling 245 families or groups of languages featured.

The most represented family is the Indo-European language family with 124 LOPs out of 867 (14.3%), however this stems primarily from LOPs used in national competitions. If we are focusing solely on LOPs from international olympiads (IOL, Asia-Pacific Linguistics Olympiad – APLO, and the Online Olympiad in Linguistics – OnLing), there are only 9 out of 130 problems (6.9%). Moreover, in the past 12 years, only two Indo-European languages were featured in these international olympiads, both of them focusing on the writing systems of extinct languages, Book Pahlavi (Nikulin, 2019) and Luwian (Lee, 2016). On the other hand, national olympiads, particularly in the beginning, are much more likely to use Indo-European languages because, on one hand, there are extensive resources and it is very easy to find native speakers of such languages who can help create a problem, and on the other hand, in any specific country, students are likely to speak or have knowledge of only a limited subset of Indo-European languages.

### *2.2. Focus on lesser-known languages*

As a consequence of the need to ensure fair conditions for participants, problems have traditionally made extensive use of lesser-known languages. In some recent contexts, particularly in national LOs, additional practical considerations have come into play. During and after the COVID-19 pandemic, many national LOs were organised online and sometimes it was impractical or even impossible to ensure live and careful invigilation. As a result, there was an increased interest in featuring languages with a small number of speakers and, particularly, with a low level of documentation and online resources. This choice also aligns with one of the main aims of the LOs, namely, to raise awareness and promote linguistic diversity. For example, the 2024 edition of the IOL's individual paper featured five problems and five distinct languages, each spoken by fewer than 2000 people: Koryak — Chukotko-Kamchatkan family, ~1665 speakers (Derzhanski, 2024), Hadza — isolate, ~1000 speakers (Ahmed, 2024), Komnzo — Yam, ~250 speakers (Davletova, 2024), Dâw — Naduhup, ~140 speakers (Oliveira Fontes, 2024), and Yanyuwa — Pama-Nyungan, 3 fluent speakers (Mirea, 2024).

### *2.3. Focus on "interesting" phenomena*

The most important advantage, particularly for the typological studies, is that LOPs tend to highlight "interesting" or non-trivial linguistic phenomena. This is not the result of neutral or random sampling, but rather the deliberate choice of problem setters, who aim to construct diverse and engaging problem sets while avoiding repetition and encouraging geographical, linguistical, and genetical variety. Particularly in the case of under-documented languages, LOPs can therefore serve as complementary information, drawing attention to less frequently described phenomena or traits. While the corpus of a LOP is not always suitable for a thorough scientific analysis, it can provide a very good starting point for future in-depth studies.

For example, Fig. 1 shows the comparison between the number of LOPs featuring different numeral systems and the number of languages with the respective systems, according to WALS (Comrie, 2013). We can easily see that while in reality most of the languages have a base-10 system, followed by those with a base 20 (or mixed 20-10 system), when it comes to a corpus of LOPs the distribution is much more uniform, with almost equal number of languages in each category. Therefore, using an LOP-based corpus gives greater visibility to less common phenomena and increase their relative frequence in the dataset for typological analysis, compared to a general typological analysis in which some uncommon phenomena may be lost, being masked by other, much more prominent, features. Moreover, general language feature databases, such as WALS, do not include rare phenomena such as overcounting or subtractive systems (Comrie, 2013).

---

[1] While linguistics problems were originally developed as pedagogical tools for teaching linguistic analysis to undergraduate students (Gleason, 1995; Merrifield, 1960), their systematic use and development as a distinct genre of problems is more closely tied to the establishment of linguistics olympiads. Earlier instances appear relatively isolated and do not yet constitute a fully established tradition and may therefore be more appropriately regarded as "proto-LOPs".

[2] "Macro language group" is used here as a broad classification label encompassing both language families and higher-level genealogical or areal groupings. As this study is not concerned with detailed linguistic taxonomy, the term is intended to avoid making any strong claims about language classification. For instance, Atlantic-Congo languages are here included under the broader Niger-Congo "macro language group".

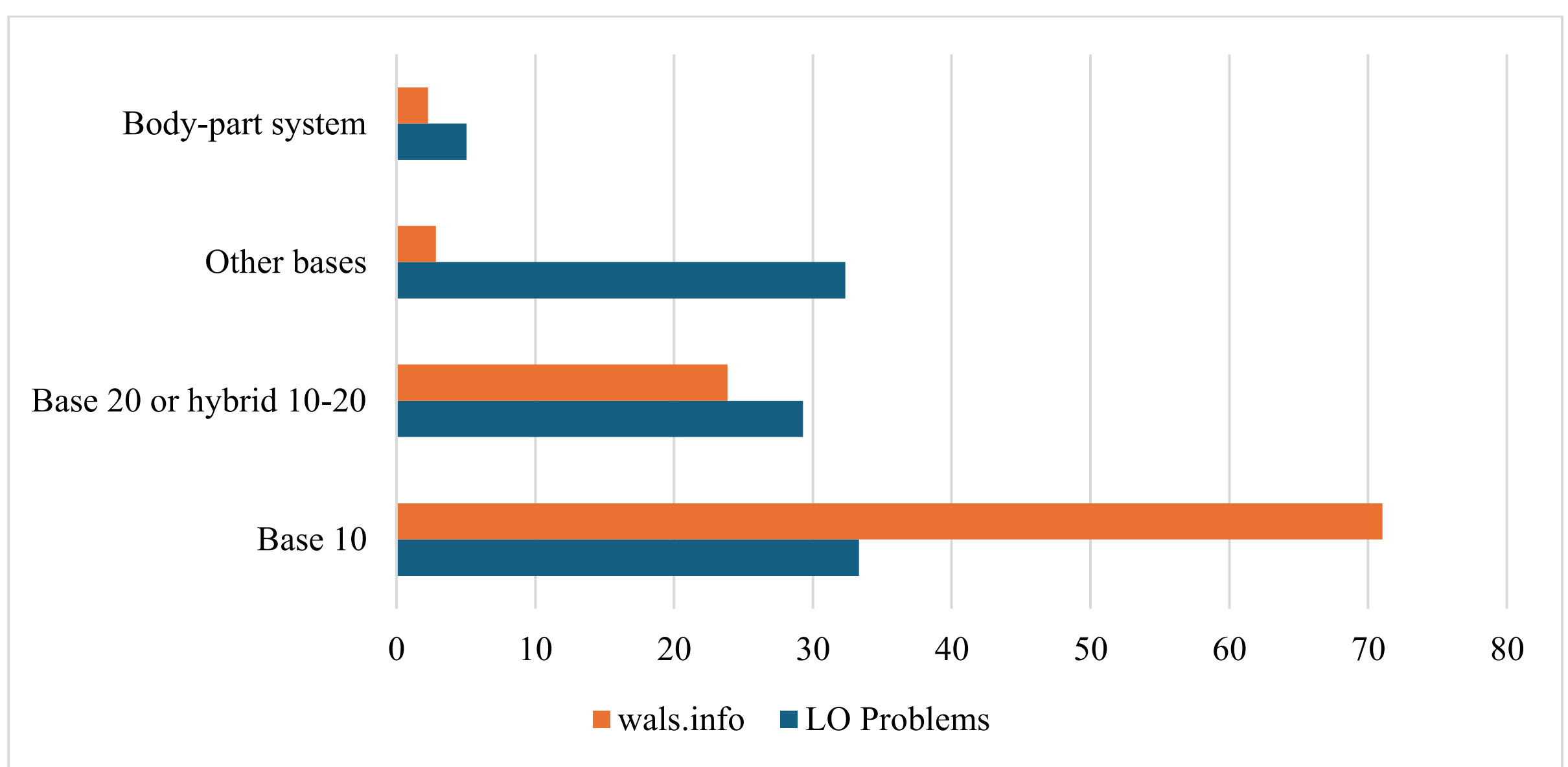

***Fig. 1.*** *Comparison of the occurrences of languages with different number systems in LOPs (blue bars) and in the WALS database.*

This perspective suggests that LOP-based research may be understood as a methodological approach that complements, rather than replaces, established areas such as linguistic typology.

### *2.4. Self-sufficiency*

Finally, an important advantage is that all LOPs are self-sufficient, meaning that the language corpus given in each problem is enough to allow anyone to solve the problem and derive a primitive grammar, based on which one can solve the tasks (Derzhanski & Payne, 2010). This property places LOPs into focus as a valuable resource for large language models (LLMs) benchmarking and understanding how LLMs process language. This idea started a few years ago, when Vaduguru *et al.* (2021) attempted to develop a synthesis model for a neural network in order to make linguistic generalisations based on a limited corpus. Since then, with the recent boom in LLMs for everyday use, more and more studies have begun focusing on LOPs as a robust way to analyse the reasoning of different artificial intelligence systems (Bean et al., 2024; Chi et al., 2024; Sánchez et al., 2024). For example, Bean *et al.* (2024) created a benchmark for advanced reasoning abilities in large language models based on the problems from the United Kingdom LO. This study aimed to evaluate the LLM capabilities to identify and generalise linguistic patterns and to follow complex task instructions. Even state-of-the-art LLMs only achieved a 38.7% accuracy, proving that solving LOPs (by operating on a minimal corpus) is complementary to large data analysis, the latter being better performed by machines.

## 3. Limitations of using LOPs in scientific research

It is also important to highlight the limitations of the LOPs in academic linguistics. A primary source of limitations is that LOPs were not originally designed for academic research. As such, their construction may prioritise self-consistency and regularity of the presented phenomena over linguistic accuracy or completeness. As a result, there could be certain simplifications or lack of information which could prove problematic for a rigorous scientific study.

At the same time, it is important to emphasise that LOPs are typically based on data drawn from existing linguistic research and studies, even if they often involve adaptation or reinterpretation for pedagogical purposes. Rather than functioning as primary sources of linguistic data, they are therefore better understood as structured and pedagogically-adapted representations of linguistic phenomena. As such, they can still serve as useful starting points for preliminary research which can then be complemented or followed by more in-depth analysis based on primary academic literature.

### 3.1. Dialectal variation

One of the key issues is that language data are not necessarily well-recorded or well-represented when it comes to LOPs, particularly the use of specific dialects. A good example is my recent problem featuring the Evenki language (Neacșu, 2024, p. 118), which features interesting vowel harmony due to three distinct but concurrent features: (1) the vowel harmony of the suffix is triggered by the first vowel of the root and not the last, thus featuring a distal vowel harmony; (2) there are three types of vowel harmony: a total assimilation, such as in the case of the plural suffix, a harmony that affects round vowels, based on backness, such as in the case of the directive-locative suffix, and a harmony based on backness that affects all vowels, as in the case of the 1st person plural possessive suffix; (3) there is a neutral vowel (<*i*>), which does not trigger, nor block vowel harmony (Kă lì nà, 2019).

While Evenki is generally considered to be spoken in Russia and most studies focus on the Russian dialects of Evenki (Atknine, 1997), there are also several dialects spoken in China (Dumont, 2017) whose grammar differs significantly from the Russian ones. The Russian dialects of Evenki, based on the work of Bulatova & Grenoble (1999) have a different vowel system than those of the Chinese dialects (Kă lì nà, 2019), as well as different vowel harmony rules. Moreover, even the vocabulary differs significantly, as shown in Table 1. Some words have completely different roots, such as in the case of ‘book’, ‘rain’, or ‘cloth’, while others have similar or even identical roots, such as ‘house’.

***Table 1.*** *Vocabulary comparison between the Russian and Chinese dialects of Evenki, preserving the original orthography.*

| Russian dialect (Bulatova & Grenoble, 1999) | Chinese dialect (Kă lì nà, 2019) | Gloss |
|---|---|---|
| *kniga* | *bitəg* | ‘book’ |
| *tigdə* | *udun* | ‘rain’ |
| *təti* | *bɵ:s* | ‘cloth’ |
| *ǯu:* | *jʉ:* | ‘house’ |

The aforementioned Evenki LOP does not mention that it is based on the Chinese dialects of Evenki and therefore it would be wrong to use it to draw conclusions about the most commonly referenced Russian dialects of the language.

### 3.2. Generally based on a single source

Another disadvantage is that certain phenomena of a certain language could be presented inconsistently throughout various sources. As an example, we can look at the recent APLO problem featuring the Kobon numeral system (Neacșu & Boroń, 2023) which is based on the reported data by Davies (2011). Based on this source, students are expected to infer the following two rules:

(1) There is a body part-based counting system for the numbers from 1 to 24, centred on 12. The numbers from 13 to 24 are formed by the rule $24 - X = X$ *böŋ daŋ*;
(2) At the same time, for higher numbers, a base 20 is used, having the scores *ñɨnjuöl* ‘20’, *ñɨnjuöl mihöp* ‘40’, and *ñɨnjuöl mɨhau nigaŋ* ‘60’. Complex numbers can be formed by adding a base number (from 1 to 23) to a score (20, 40, or 60) based on the rule $X$ *ado gɨ da* $Y = X + Y$, where $X$ represents the scores and $Y$ represents the units (from 1 to 23).

Nevertheless, a previous study from Davies (1984) assigns the values of the scores to 23, 46, and 69, respectively, thus shifting to a base-23 system. Finally, in an even earlier study on Kobon (Dawson & Dawson, 1974), the authors mention that “No attempt is here made to present a formula for the numeral phrase as to do so is considered too complicated and impractical.”

It is therefore important to notice that LOPs may rely on a single source, which is not necessarily the most accurate when featuring a particular phenomenon.

### *3.3. Reduced paradigms*

While the first two disadvantages are somehow extrinsic and depend on the sources used to create the problem and/or the description of the language featured in the problem, in the next two subsections I will focus on two intrinsic disadvantages, stemming primarily from the way that each author chose to construct the LOP, bearing in mind that its main purpose is to be used in a pre-university academic competition.

Sometimes, a complete paradigm of a phenomenon is too complex to be represented by a corpus with a relatively small size, as is the case in most LOPs. A relevant example is the verb morphology of the Yimas language (Neacșu & Krashtan, 2023). In reality, the Yimas verb morphology is extremely complex with the basic grammar of Yimas extending the marking of pronominal affixes for core nominals on over 35 pages (Foley, 1991). The problem presented the paradigm of person marking in intransitive and transitive sentences in the affirmative perfective and in the case where a modal marker was added, representing either the negative or the potential modality (translated into English using the adverb 'almost'). The complete paradigm for the Yimas verb structure is shown in Table 2.

***Table 2.*** *Yimas verb structure, based on transitivity and on the presence of modal particles. S and O refer to the subject and object, respectively. ns = non-singular* (adapted from Foley (1991)).

| | Intransitive | Transitive |
|---|---|---|
| No modal particle (affirmative) | $S - V - t$ | $O - S - V - t$<br>(if O = 3)<br>$S - kampan - V - t$<br>(if S = 1 and O = 2sg)<br>$S - O - V - t$<br>(if S = 3 and O = 1 or 2; if S = 2 and O = 1; if S = 1 and O = 2ns) |
| Modal particle ($MOD$), either negative (*ta*) or potential (*ant*) | $MOD - S - V - r/c - Num$<br>(if S = 1 or 2ns)<br>$MOD - pun - V - t$<br>(if S = 2sg)<br>$MOD - pun - V - r/c - Num$<br>(if S = 3) | $MOD - S - V - r/c - Num$<br>(if O = 3)<br>$MOD - O - V - r/c - Num$<br>(if S = 3) |

This verb structure is extremely complex, notwithstanding each subject marker having two forms depending on whether it is placed word-initially or word-medially. As such, asking students to deduce this full paradigm based on fewer than 25 examples is not feasible.

The actual problem reduced the paradigms in two ways: (1) it avoided the 2nd person singular and (2) it avoided the 1st person subject. Excluding these cases, the simplified paradigm is shown in Table 3. This reduced paradigm can also be reinterpreted based on the presence of 3rd singular subject or object, based solely on the role of the third person argument.

***Table 3.*** *Reduced Yimas verb structure, based on transitivity and on the presence of modal particles. S and O refer to the subject and object, respectively. ns = non-singular* (adapted from Neacșu and Krashtan (2023)).

| | Intransitive | Transitive |
|---|---|---|
| No modal particle (affirmative) | $S - V - t$ | $O - S - V - t$<br>(if O = 3)<br>$S - O - V - t$<br>(if S = 3 or O ≠ 3) |
| Modal particle ($MOD$), either negative (*ta*) or potential (*ant*) | $MOD - S - V - r/c - Num$<br>(if S ≠ 3)<br>$MOD - pun - V - r/c - Num$<br>(if S = 3) | $MOD - S - V - r/c - Num$<br>(if O = 3)<br>$MOD - O - V - r/c - Num$<br>(if S = 3) |

It is therefore important to note that while the problem presents the complex verb morphology of Yimas, highlighting a distinction between word-initial and word-medial subject markers, different structures based on transitivity, presence of modal markers, and the third person argument, in reality the paradigm is much more complex, also including a distinction between the 2nd person singular and non-singular.

Another example of reduced paradigms is Koryak (Derzhanski, 2024), which only includes a subset of agent and patient combination that is internally consistent and excludes combinations that are irregular or unpredictable.

### *3.4. Cherry-picking*

This is perhaps the most dangerous disadvantage in using LOPs. It is, to some extent, similar to the reduction of paradigm, but does not operate consistently throughout the paradigm. In other words, the author chooses those examples which particularly align with the rules they want to feature or demonstrate.

On one hand, we can talk about exceptions that diverge from the regular paradigm, and which are universally present in the world's language. Most LOPs' authors purposefully avoid these exceptions and only focus on the regular paradigms. In some cases, the authors actually include some exceptions either in the corpus or in the tasks, in order to show students that languages are not always as regular as one might think. For example, a problem dealing with the vowel harmony in Turkish (Bozhanov, 2010) included in the corpus two words that do not follow the rules of vowel harmony because their roots were loaned from another language (*ikbal* 'prosperity' and *takat* 'strength', both of them borrowed from Arabic).

While avoiding general exceptions to the paradigm is something mostly harmless, the danger arises when the author of a problem specifically selects certain words or stems in order to represent a rule that is not actually attested in the language.

Of course, besides these disadvantages, there could also be issues with the problem composition, including problems that are inconsistent, not self-sufficient, or containing parasitic solutions, i.e., alternative ways to explain the corpus yielding completely different results. A further source of potential distortion arises from the use of a metalanguage accessible to solvers, which may lead to the reinterpretation or reduction of linguistic distinctions (for example, mapping aspectual distinctions onto temporal categories or treating paucal as plural) and thus to partial loss of information.

For this reason, any academic research should be conducted on carefully selected collections of LOPs, curated by experienced problem setters or researchers familiar with this genre of problems.

## 4. Connections between LOPs and academic linguistics fields

So far, I have presented some advantages and disadvantages of using LOPs in academic research, but in order to make the implications of this research more tangible, it is important to also outline a few subfields of linguistics in which LOPs could be beneficial.

LOPs are directly connected first and foremost to linguistic typology, since they aim to highlight language diversity and the different mechanisms that languages may employ to express certain features. One such example was used in the Romanian LO (Neacșu, 2016), in which solvers were given four sentences translated into four different languages, each language featuring a different morpho-syntactic alignment (Dyirbal – ergative-absolutive, Nez-Perce – tripartite alignment, Tahitian – direct alignment, and Wappo – nominative-accusative alignment). This highlighted the diverse ways in which different languages mark the core arguments of transitive and intransitive sentences.

Secondly, these problems are also intertwined with linguistic relativity, having featured phenomena such as colour perception (Neacșu, 2018) based on the study of Berlin and Kay (1969), space perception (Littell, 2018), and kinship terminology (Derzhanski, 2014; Gilyarova, 2018).

Natural language processing is also closely related to solving LOPs, since the thought process is similar to data interpretation by machine translators. Both involve inferring structure and meaning from limited linguistic data. In particular, recent work has used LOPs as tools for "in-context identification and generalisation of linguistic patterns in very low-resource or extinct languages" (Bean et al., 2024). From this perspective, the thought process of solving such problems is similar to machine translation involving low-resource languages, where mappings between linguistic forms must be inferred from small datasets rather than extracted from large corpora. Recent works explicitly describe LOP solving as combining reasoning with translation-like capabilities in the absence of extensive background knowledge (Bean et al., 2024; Lin et al., 2025). At the same time, LOPs are complementary to standard machine translation. Unlike most translation systems, which rely on large parallel corpora, LOPs operate on extremely limited corpora/datasets. Consequently, their solution depends not only on pattern induction, but also on the solver's understanding of their own native language in order to interpret and generalise from the given data.

Finally, the process of creating LOPs, as opposed to solving them, is closely linked to linguistic fieldwork. Both cases involve working with limited data to select or construct examples that give enough information to allow reliable generalisations. In linguistic fieldwork, researchers often rely on constrained interaction with informants and must carefully elicit data that reveals relevant structural properties of the language. Similarly, when constructing a LOP, problem setters aim to select and organise examples such that all necessary patterns or phenomena are discoverable or inferable from the given data. In other words, one needs to ensure that the data provided are both internally consistent and sufficiently rich to allow the intended generalisations to be made.

For example, the morphosyntactic alignment can be primarily determined only through three examples (a-c) in which the analysed word is featured as a subject of an intransitive verb (a), a subject of a transitive verb (b), and an object (c). This is exemplified below for four languages with four different morphosyntactic alignments[3]. Of course, the number of examples could be reduced to two if the same noun were used as both the subject and the object of a single transitive sentence. However, this can potentially generate unnatural sentences that are unlikely to occur in unelicited data.

1. Mandarin Chinese, direct alignment (constructed examples)

(1a) *儿子* *休息。*
boy rest
'The boy rests.'

(1b) *儿子* *看见* *苹果。*
boy see apple
'The boy sees the apple.'

(1c) *女儿* *看见* *儿子。*
girl see boy
'The girl sees the boy.'

2. Russian, nominative-accusative alignment (constructed examples)

(2a) *Мальчик-Ø* *отдыхал.*
boy-NOM rest.3SG.PST
'The boy rested.'

(2b) *Мальчик-Ø* *увидел* *вор-а.*
boy-NOM see.3SG.PST thief-ACC
'The boy saw the thief.'

(2c) *Отец-Ø* *увидел* *мальчик-а.*
father-NOM see.3SG.PST boy-ACC
'The father saw the boy.'

3. Dyirbal, ergative-absolutive alignment (example 3a adapted from Dixon (1972) examples 3b-3c from Neacșu (2016))

(3a) *Yaṛa* *baninyu.*
man.ABS come.PST
'The man came.'

[3] A list of abbreviations used in glossing is provided here: 3>3 = subject and object both in the third person, 3SUBJ = 3rd person subject in intransitive clauses, ABS = absolutive, ACC = accusative, ERG = ergative, INTR = intransitive, NOM = nominative, PST = past, SG = singular.

| | | | |
|---|---|---|---|
| (3b) | *Jugumbil* | *yaṛa-ŋgu* | *buṛan.* |
| | woman.ABS | man-ERG | see.PST |
| | 'The man saw the woman.' | | |
| (3c) | *Yaṛa* | *jugumbil-ŋgu* | *buṛan.* |
| | man.ABS | woman-ERG | see.PST |
| | 'The woman saw the man.' | | |

4. Nez-Perce, tripartite alignment (adapted from Neacşu (2016))

| | | | |
|---|---|---|---|
| (4a) | *Háma* | *hixne.* | |
| | man.INTR | 3SUBJ:see.PST | |
| | 'The man saw.' | | |
| (4b) | *Háma-nim* | *peexne* | *'áyat-ne.* |
| | man-ERG | 3>3:see.PST | girl-ACC |
| | 'The man saw the girl.' | | |
| (4c) | *'áyat-nim* | *peexne* | *háma-ne.* |
| | girl-ERG | 3>3:see.PST | man-ACC |
| | 'The girl saw the man.' | | |

It is important to mention that this approach would not take into account split alignments, in which two (or more) basic morphosyntactic alignments co-occur in a complementary distribution based on grammatical features (such as Pashto (Indo-European), which uses ergative-absolutive alignment in past tense, but nominative-accusative in present tense), nominal categories (Pitjantjatjara (Pama-Nyungan) uses nominative-accusative for pronouns, but ergative-absolutive for nouns), or even the number of the pronoun arguments (Ngiyamba (Pama-Nyungan) uses nominative-accusative alignment for the first and second persons, but ergative for the third). Similarly, this approach should be further improved to cover non-classic morphosyntactic alignments such as the split intransitive (with the two subtypes: fluid-S and split-S) or the direct-inverse (pronoun hierarchy).

## 5. Conclusions

This manuscript opens the road to include LOPs in academic research by highlighting a number of the advantages that make LOPs a unique tool in academic research, while also addressing their limitations and how these may be mitigated. Self-consistency, a key attribute of LOPs, makes them particularly suitable candidates for applications such as LLM benchmarking. Moreover, the LOPs' focus on linguistic rara makes them a complementary corpus for typological research, as it increases the relative frequency of such phenomena within a curated dataset.

At the same time, I discuss the potential drawbacks of using LOPs in academic research, including potential incomplete information on the language data, issues related to dialectal variation, paradigm reduction, and the influence of problem construction choices. These factors highlight the importance of careful selection and critical use of LOP-based corpora in academic research.

Finally, several avenues for future research have been outlined. With the growing popularity of linguistics olympiads and the ever-growing numbers of countries and students participating, it becomes increasingly relevant to find a place for LOP-based research within academic linguistics.

In this respect, LOP-based research is best understood not as a subfield of linguistic typology or computational linguistics per se, but rather as a cross-cutting methodological framework, which operates at the intersection of linguistic analysis, pedagogy, and data curation, providing structured and deliberately curated representations of linguistic phenomena. While the selection of LOPs is not governed by typological sampling principles, the resulting datasets may nevertheless complement typological research by highlighting a broad range of non-trivial and less frequently described phenomena. As such, LOP-based research occupies a methodological niche, contributing tools and perspectives that can support, but not replace, established approaches in linguistic typology and related fields.